\documentclass[journal, web ,twopage]{ieeecolor}

\usepackage{url}
\usepackage{graphicx}
\usepackage{hyperref}
\usepackage{generic}

\usepackage{amsmath}
\usepackage{graphicx}
\usepackage{float}
\usepackage{stmaryrd}
\usepackage{amssymb}
\usepackage{dsfont}

\usepackage{bbold}
\usepackage[noend]{algpseudocode}
\usepackage{algorithm}
\let\labelindent\relax
\usepackage[outdir=./]{epstopdf}

\newcommand\Algphasebegin[1]{%
\vspace*{-.7\baselineskip}\Statex\hspace*{\dimexpr-\algorithmicindent-2pt\relax}\hspace{5mm}%
\Statex\hspace*{-\algorithmicindent}\textbf{#1}%
\vspace*{-.7\baselineskip}\Statex\hspace*{\dimexpr-\algorithmicindent-2pt\relax}\hspace{3mm}\rule{0.485\textwidth}{0.7pt}%
}
\newcommand\Algphaseinside[1]{%
\vspace*{-.7\baselineskip}\Statex\hspace*{\dimexpr-\algorithmicindent-2pt\relax}\hspace{3mm}\rule{0.48\textwidth}{0.7pt}%
\Statex\hspace*{-\algorithmicindent}\textbf{#1}%
\vspace*{-.7\baselineskip}\Statex\hspace*{\dimexpr-\algorithmicindent-2pt\relax}\hspace{3mm}\rule{0.48\textwidth}{0.7pt}%
}

\newcommand\Algphaseunsupervised[1]{%
\vspace*{-.73\baselineskip}\Statex\hspace*{\dimexpr-\algorithmicindent-2pt\relax}\hspace{3mm}%
\Statex\hspace*{-\algorithmicindent}\textbf{#1}%
\vspace*{-.73\baselineskip}\Statex\hspace*{\dimexpr-\algorithmicindent-2pt\relax}\hspace{3mm}%
}

\usepackage{enumitem}
\usepackage{textcomp}
\def\BibTeX{{\rm B\kern-.05em{\sc i\kern-.025em b}\kern-.08em
    T\kern-.1667em\lower.7ex\hbox{E}\kern-.125emX}}
\begin{document}

\title{Deep ContourFlow: Advancing Active Contours with Deep Learning}
\author{Antoine Habis \IEEEmembership{Member, IEEE}, Vannary Meas-Yedid \IEEEmembership{Member, IEEE}, Elsa Angelini \IEEEmembership{Member, IEEE}, and Jean-Christophe Olivo-Marin \IEEEmembership{Fellow, IEEE}
\thanks{Manuscript received. This project has received funding from the Innovative Medicines Initiative 2 Joint Undertaking under grant agreement No 945358. This Joint Undertaking receives support from the European Union’s Horizon 2020 research and innovation program and EFPIA. www.imi.europe.eu. The project was also partially funded through the PIA INCEPTION program (ANR-16-CONV-0005).}
\thanks{A. Habis, V. Meas-Yedid and J.-C. Olivo-Marin are with the Bioimage Analysis Unit, Institut Pasteur, CNRS UMR 3691, Université Paris Cité, 75015 Paris, France (e-mail: antoine.habis@pasteur.fr; vannary.meas-yedid-hardy@pasteur.fr; jcolivo@pasteur.fr)  }
\thanks{A. Habis and E. Angelini are with LTCI, Télécom Paris, Institut Polytechnique de Paris, France (e-mail: elsa.angelini@telecom-paris.fr)}}

\maketitle

\begin{abstract}
 This paper introduces a novel approach that combines unsupervised active contour models with deep learning for robust and adaptive image segmentation. Indeed, traditional active contours, provide a flexible framework for contour evolution and learning offers the capacity to learn intricate features and patterns directly from raw data. Our proposed methodology leverages the strengths of both paradigms, presenting a framework for both unsupervised and one-shot approaches for image segmentation. It is capable of capturing complex object boundaries without the need for extensive labeled training data. This is particularly required in histology, a field facing a significant shortage of annotations due to the challenging and time-consuming nature of the annotation process. We illustrate and compare our results to state of the art methods on a histology dataset and show significant improvements. 
 
 \end{abstract}


\section{Introduction}
\label{sec:introduction}
\IEEEPARstart{I}NTEGRATION of deep learning techniques into histological image analysis has emerged as a powerful tool, revolutionizing the field of pathology. Traditional methods of histological analysis have often relied on manual interpretation, which is subjective, and prone to inter-observer variability. The advent of deep learning and  particularly convolutional neural networks (CNNs), has paved the way for more accurate, efficient, and reproducible segmentation of histological images. However, a significant challenge lies in the extensive need for large annotated datasets that deep learning models require for optimal performance. However, those are not always available in histological studies due to the numerous different type of tissues, the intricate nature of tissue structures and the labor-intensive process of manual annotation. 

Here, we propose two methods based on unsupervised and one-shot learning approaches to evolve active contours by efficiently leveraging minimal annotated samples and grasp the complex patterns inherent in natural histological images. The first is completely unsupervised and is used to segment complex objects in front of complex textures. The second is a one-shot learning segmentation approach and is used  for instance-based segmentation of dilated tubules in the kidney, which is an important challenge in histology. Indeed, dilation of renal tubules can occur in various pathological conditions and may be a sign of an underlying issue such as obstruction in the urinary tract, an inflammation of the interstitial tissue in the kidney, congenital anomalies, or a Polycystic Kidney Disease (PKD). It is therefore essential to be able to automatically detect these potential diseases at a sufficiently early stage. 

The article is divided into two methodological sections. \autoref{unsupervised} describes its unsupervised version and \autoref{oneshot} its declination as a one-shot learning algorithm. To our knowledge, unlike \cite{DEEP_SNAKE,POLY_RNN,POLY_RNN++, END}, this is the first paper to use the relevance of CNN's features while retaining the spirit of active contours that require no training. Our method introduces two differentiable functions that turn a contour into a mask or a distance map in order to select effective and relevant features learned by a pre-trained neural network at several scales and move the contour towards a preferred direction using gradient descent. These two functions are coded as layers in the python torch-contour library we have provided and the implementation of the algorithms is available at \url{https://github.com/antoinehabis/Deep-ContourFlow}.

\section{Related work}
\label{related_work}

Before the 2000s, a wide range of contributions focused on unsupervised segmentation. Unsupervised segmentation offers great versatility and is therefore applicable across diverse domains, including medical imaging, remote sensing, and industrial applications, where acquiring labeled data is often resource-intensive.

Among all of these methods, the main contributions involve energy minimization such as thresholding techniques \cite{OTSU}, active contours \cite{SNAKE, GEODESIC, GVF, ACWE, JC, JC2} and graph cut models \cite{GRAPHCUT, GRABCUT}. These methods, and particularly active contours, constitute a major inspiration behind the creation of our own model.  

In the 2010's, the advent of deep learning and particularly convolutional neural networks (CNNs) \cite{ALEXNET, VGG16, INCEPTION, GOOGLE_NET, RESNET}, played a crucial role in advancing unsupervised segmentation techniques. These models can autonomously learn hierarchical representations from raw data, allowing them to capture intricate patterns and features within images. When we think of deep learning, it is often assumed that an abundant amount of data for training purposes is needed to ensure good performances. However, this is not necessarily the case. For instance, some methods take advantage of existing trained models to perform clustering in the features space \cite{USCF, USCF2, STYLEGAN_unsupervised}. Others use only the extremely convenient architecture of CNNs, combined with an a-priori knowledge on the objects of interest, to perform segmentation \cite{VAE}. Even so, unsupervised segmentation remains relatively limited, especially when the objects of interest in the image have highly complex properties. Therefore, in several cases additional information is required.

One-shot learning uses a single instance of each object to be segmented to perform the segmentation. Most one-shot learning segmentation algorithms work by having a query image and an annotated support image with a given mask. The aim is to find the object delineated in the support image within the query image.

This is often achieved by using a similarity measure between the features of the query and those of the support in order to find the best query mask that maximizes the similarity. 

As mentioned in \cite{REVIEW}, one-shot learning algorithms for semantic segmentation can be divided into three broad overlapping categories: 

\begin{itemize}
\item Prototypical networks \cite{PFENET, SGONE, FSSPL, SMRCNN} work by passing the support image through a feature extractor and averaging the features of the class to be segmented to obtain a representation of the class in the latent space. This representation is then used to classify the pixels in the query image as belonging or not to the class using a similarity with the extracted features of the query.
\item Conditional Networks \cite{OSLSM, CNFSS} use a different approach. They decouple the task into two sub-tasks, each representing a branch in the neural network. The main branch extracts features from the image query by projecting them into a well-defined space. The secondary branch takes as input the support image and selects the features of the region of interest using its associated mask. Then, a conditioning convolutional branch assists the reconstruction to return the predicted query mask.

\item Finally, latent space optimization algorithms \cite{GANORCON, DATASETGAN} seek to refine the latent representation of the image using generative adversarial networks (GAN) or variational autoencoder (VAE), and then use this representation to segment the class present in the support image within the image query. 
\end{itemize}

In the above methods, a neural network is always trained beforehand with a dataset containing one support image and its corresponding mask for each of the classes. 
Our one-shot learning method falls into the category of prototypical networks, but unlike the methods mentioned above, no additional training is required. Our proposed method uses only the weights of VGG16 \cite{VGG16} trained on ImageNet \cite{IMAGENET} and extracts the features directly for segmentation with a fit and predict step. Before presenting the one-shot version of our algorithm in \autoref{oneshot}, we start by introducing its unsupervised version.

\section{Method: Unsupervised learning}
\label{unsupervised}
\subsection{Context}

Our proposed framework that we coin Deep ContourFlow (DCF) aims at iteratively evolving a contour $C \in [0,1]^{n_{nodes} \times 2}$ by using the features derived from image $I \in [0,1]^{H \times W \times 3}$ through the pre-trained VGG16 \cite{VGG16} architecture across various scales. This concept is inspired by the Chan-Vese algorithm \cite{ACWE} (CV), it can work on any type of images including colour images but uses averages of complex deep learning features at multiple scales rather than simple averages of grayscale intensity values like in CV.

The VGG16 \cite{VGG16} encoder encompasses five scales, where $f_s$ denotes the layer's output preceding the $s^{th}$ max pooling, diminishing spatial dimensions by a factor of $2^s$. The set $\left\{f_s, s \in \mathcal{S} = \left\{ 0,..,4 \right\} \right\}$  encapsulates the multi-scale encoding of $I$. In an entirely unsupervised framework, a straightforward approach is to find the contour $C$ that maximizes the difference between features inside and outside of it. To implement this, we need to build a function generating a mask from the specified contour, facilitating the selection of these two feature sets. Furthermore, for iterative contour adjustments using gradient descent, this function must be differentiable.

\subsection{Fully differentiable contour{-}to{-}mask mapping}
\label{mask}

\begin{figure}[!htb]
\centerline{\includegraphics[width = 9cm]{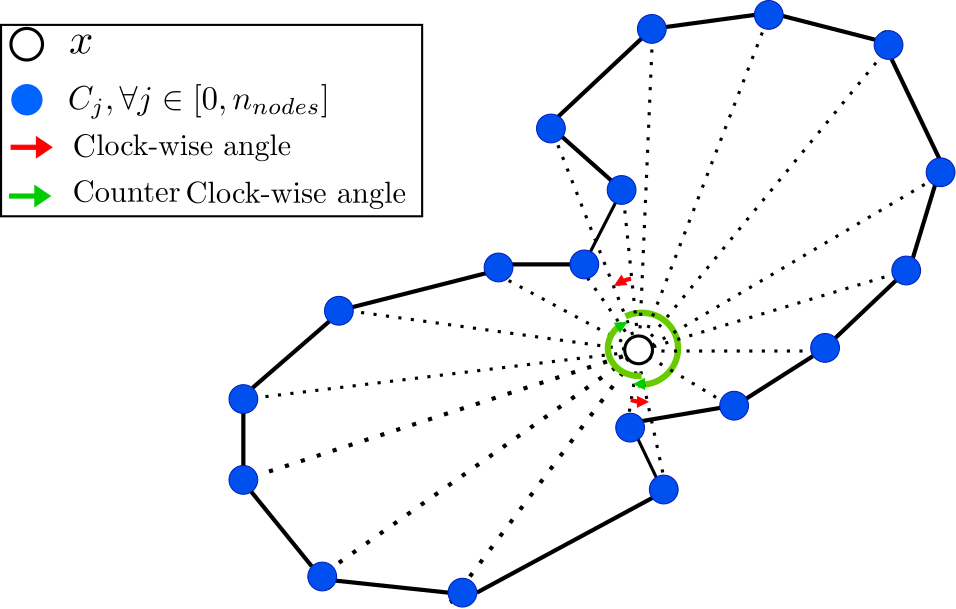}}
  \caption{Example of a polygon and representation of the oriented angles connecting an interior point $x$ to its nodes $C_j$ along the contour.}
 \label{oriented_angles}
\end{figure}
\textcolor{black}{
Let $\Omega =  \llbracket 0,H\rrbracket \times \llbracket 0,W \rrbracket$ be the index grid corresponding to the coordinates of each pixel in $I$.
The first step is to build a fully differentiable function that turns the deformable contour $C$ into a binary mask $M$. For this, we use the property that for any point $x$ inside $C$ the sum of the oriented angles is $2\pi$ and quickly converges to $0$ outside of $C$ (see \autoref{oriented_angles}). Thus we define $M$ as:}

\begin{equation}
\label{contour_to_mask}
    \forall x \in \Omega,\quad  M(x) = F_{cm}(x,C)
\end{equation}
with,
\begin{center}

    $F_{cm}(x,C) = \frac{1}{2\pi} \sum_{i = 0}^{n_{nodes}-1} \overbrace{\underbrace{u(C_i,C_{i+1},x)}_\text{orientation}\times \underbrace{\theta(C_i,C_{i+1},x)}_\text{angle}}^\text{oriented angle}$
\end{center}
\begin{align*}
     u(C_i,C_{i+1},x) &= tan_h(k \times(C_i - x)\wedge (C_{i+1} - x)) \\
     \theta(C_i,C_{i+1},x) &= arccos(\frac{<C_i - x, C_{i+1} - x>}{||C_i - x|| ||C_{i+1} - x||})
\end{align*}
 where $C_{i}$ is the $i^{th}$ node of $C$. As it is a closed contour we force the equality $C_0 = C_{n_{nodes}}$. $u$ measures the orientation of each angle ($u\approx 1$ for counter clock-wise orientation and $u\approx -1$ for the opposite direction (see \autoref{oriented_angles}) and $\theta$ the corresponding angle value. $k$ is a control parameter which must be chosen beforehand to ensure a good trade-of between an accurate estimation of $sign(\cdot)$ and a smooth differentiable function.

The mask $M_s$ at scale $s$ is then obtained by downscaling $M$ using bilinear interpolation at scale $s$ resulting in a mask with spatial size $\frac{H}{2^s} \times \frac{W}{2^s}$.

\begin{figure*}[!htbp]
    \centering
    \noindent\makebox[\textwidth]{\includegraphics[width=\textwidth]{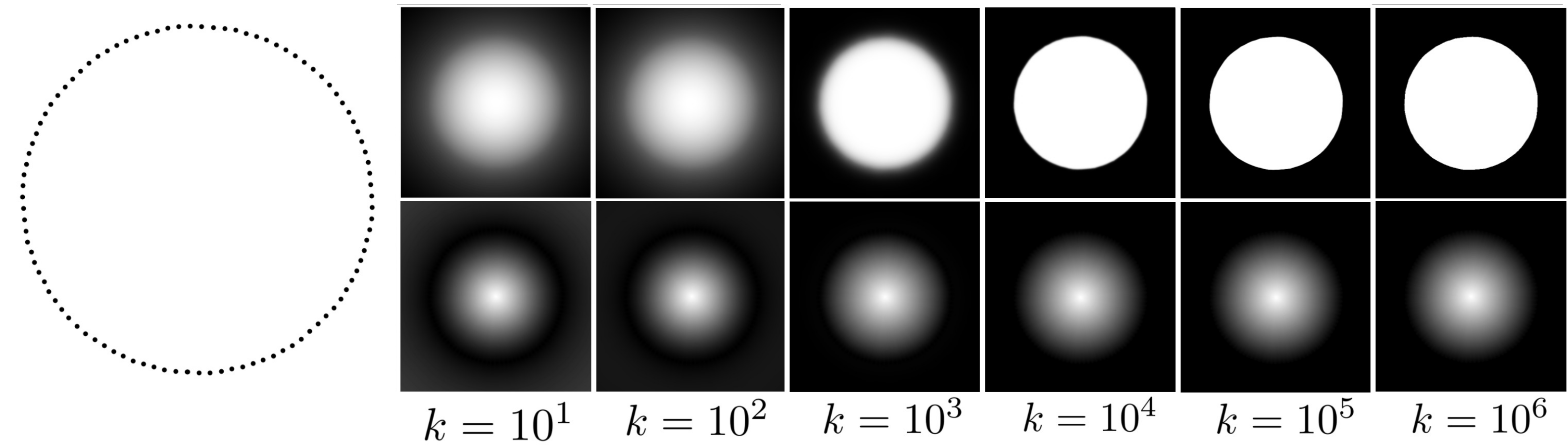}}
    \caption{Approximation of the mask of a circle using $F_{cm}$ (top row) and the distance map using $F_{cd}$ (bottom row) when increasing k with $n_{nodes}=100$.}
    \label{increasing_k}
\end{figure*}

\begin{figure*}[!htbp]
    \centering
    \noindent\makebox[\textwidth]{\includegraphics[width=\textwidth]{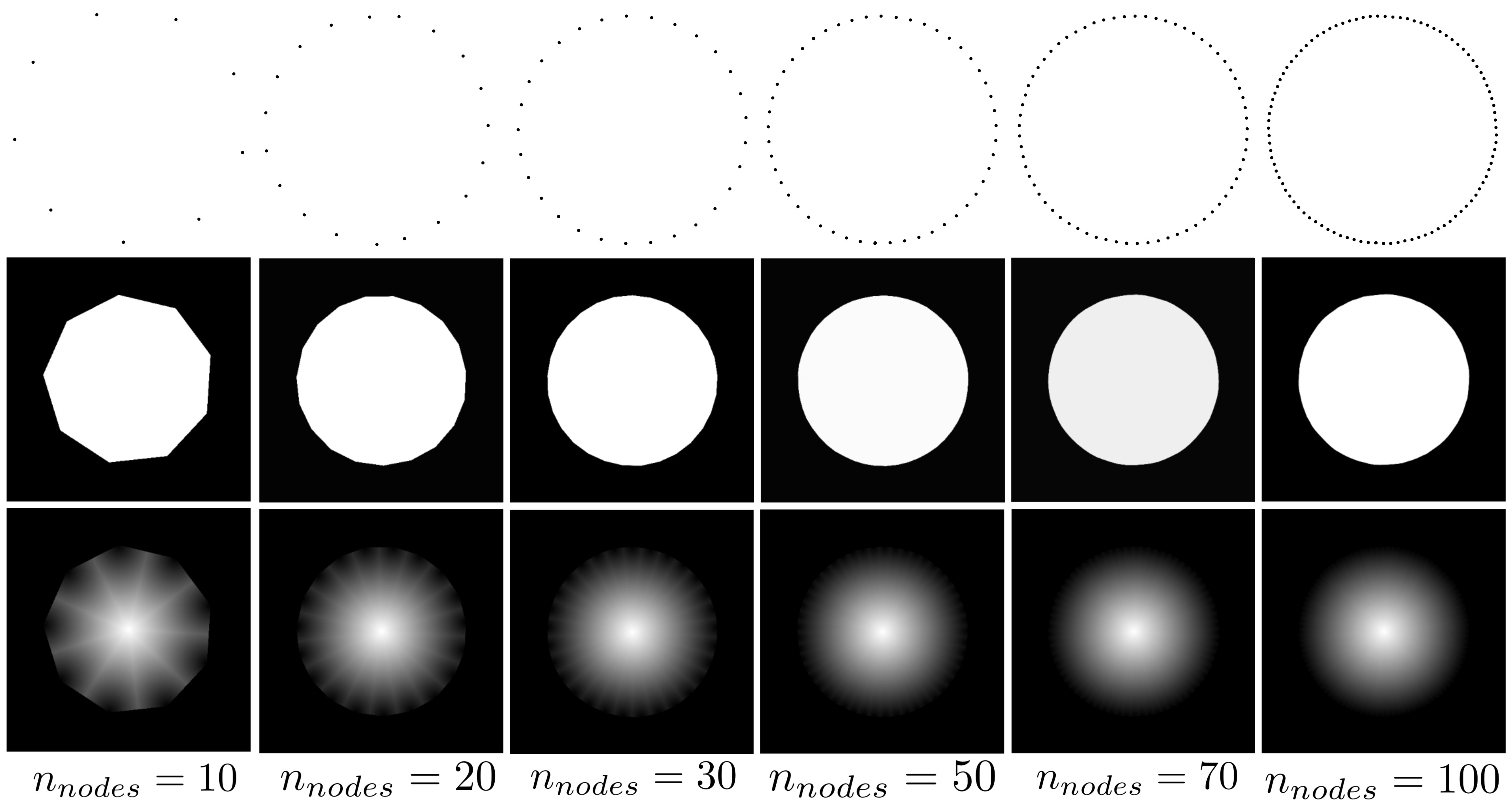}}
    \caption{Approximation of the mask of a circle using $F_{cm}$ (middle row) and the distance map using $F_{cd}$ (bottom row) when increasing $n_{nodes}$ with $k=10^{5}$.}
    \label{increasing_nnodes}
\end{figure*}

As can be seen in \autoref{increasing_k} (top row), if we take the basic example of a circle we obtain an accurate approximation of the mask using $F_{cm}$  with sufficiently large values of k (around $10^{5}$). For the approximation to be correct, we also need a sufficiently large number of points. The more points used to describe the object, the more accurate the result. The results of the mask obtained by varying $n_{nodes}$ between $10$ and $100$ with $k=10^{5}$ can be seen in \autoref{increasing_nnodes} (top row).

\subsection{Evolution of the contour}

To initiate the contour evolution process, we need to define an inital contour, denoted as $C = C^0$.

Having established a differentiable function for contour-to-mask mapping, we can now derive $[f_s]^{in}$ and $[f_s]^{out}$, representing the respective averages of feature values within and outside $C$ at each scale $s$:

 \begin{equation}
\label{features_inside_outside}
    \begin{split}
        [f_s]^{in} &=L(M_s, f_s)\\
        [f_s]^{out} &= L(1 - M_s, f_s)\\
    \end{split}
\end{equation}
with,

 \begin{equation}
    \begin{split}
        L(M_s, f_s) & = \frac{\sum_{x,y \in \Omega_s} M_s(x,y) \times f_s(x,y)}{\sum_{x,y \in \Omega_s} M_s(x,y)}
    \end{split}
\label{definition_L}
\end{equation}
and $\Omega_s  = \llbracket 0,...\frac{H}{2^s} \rrbracket \times \llbracket 0,..,\frac{W}{2^s}\rrbracket$.

\begin{algorithm}[htb]
\caption{Deep ContourFlow (Unsupervised) }
\label{dac_unsupervised}
\begin{algorithmic}
\Algphaseunsupervised{}
\textbf{Input}: $I$, $C^0$
\textbf{Output}: $C$

\vspace{1mm}
\State{$C \gets C^0$}
\State{$\Omega \gets (i,j)_{i \in \llbracket 0,H\rrbracket, j \in \llbracket 0,W\rrbracket}$}
\State{$grad = +\infty$}
\State{$\left\{f_s\right\}_{s \in \mathcal{S}} \gets \text{VGG16}(I)$}

\For{$j = 1,...,n_{epochs} $}
    \If{$|grad| > t$}
        \State{$M \gets F_{cm}(\Omega,C)$(\ref{contour_to_mask})}
        \For{$s \in \mathcal{S}$}
            \State{$M_s \gets Downscale(M,scale = s)$ (\ref{mask})}
            \State{$[f_s]^{in} \gets L(M_s, f_s)$ (\ref{features_inside_outside})}
            \State{$[f_s]^{out} \gets L(1 - M_s, f_s)$ (\ref{features_inside_outside})}
        \EndFor
    \State{$Loss \gets loss_1([f]^{in}, [f]^{out})$ (\ref{loss unsupervised})}
    \State{$grad = \nabla_C Loss$}
    \State{$C \gets C - l_{r} \times \text{Clip}(grad)$}
    \State{$C \gets$ Clean$(C)$}
    \State{$C \gets \text{Interp}(C)$}
    \EndIf
    \State{\textbf{else}{~Break$;$}}
\EndFor
\Return{$C$}
\end{algorithmic}
\end{algorithm}

Algorithm \ref{dac_unsupervised} summarises all the steps involved in moving $C$ with gradient descent.

\subsection{Loss function and contour adjustment operators}

The loss function used to evolve the contour is the following:

\begin{equation}
 loss_1([f]^{in}, [f]^{out}) = - \sum_{s \in \mathcal{S}}\frac{1}{2^s}\frac{||[f_s]^{in} - [f_s]^{out}||_2}{||f_s||_2}
 \label{loss unsupervised}
\end{equation}

$loss_1$ is a variant of the Mumford-Shah functional \cite{MUMFURD}. Whereas the latter aims to make the features inside and outside the contour homogeneous, the former aims to keep the features on the inside distant from the features on the outside. 

\begin{figure}[!ht]
  \centerline{\includegraphics[width = \columnwidth]{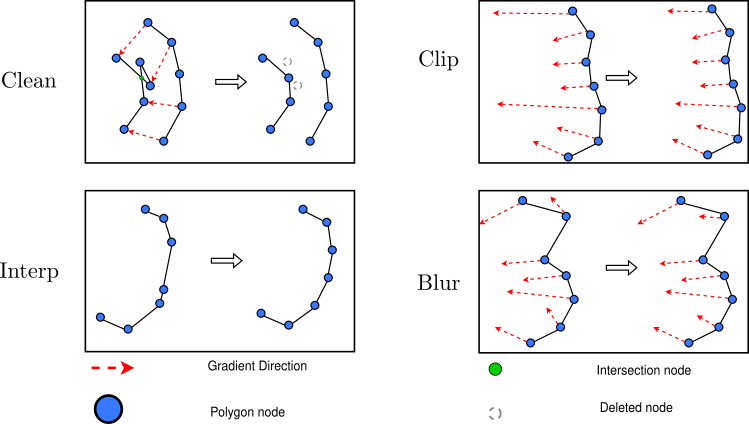}}
  \caption{Illustration of the effects of the operators used in the iteration adjustments. From top to bottom and left to right:  Clean, Clip, Interp and Blur operators are used respectively to delete loops, impose a maximum displacement, resample the points along the contour and regularize the displacement.}
 \label{iterations_adjustments_images}
\end{figure}

We remove potential loops along the contour using the Clean operator (\autoref{iterations_adjustments_images}: Top left) after each gradient descent by using the Bentley-Ottman \cite{BENTLEY} algorithm to increase time efficiency. Instead of checking all the possible intersections with each edges of the polygon, the Bentley-Ottman \cite{BENTLEY} algorithm finds $k$ intersections between $n$ segments with a time complexity of $\mathcal{O}((n+k)\log{}n)$. We also clip the norm of the gradient (\autoref{iterations_adjustments_images}: Top right). This helps gradient descent to have reasonable behavior even if the loss landscape of the unsupervised DCF is irregular.

Finally, we interpolate the contour with the Interp operator so that the distance between each consecutive point stays the same (\autoref{iterations_adjustments_images}: bottom left).

We also add a constraining force $\mathcal{F}_{\mathcal{A}rea}$ to the loss, as in the classical theory of active contours to avoid that the contour collapses:

\begin{equation}
    \mathcal{F}_{\mathcal{A}rea}(C) =  - \lambda_{area} \times \mathcal{A}rea(C)
\label{area}
\end{equation}

\subsection{Parameters and results}

\begin{figure*}[!htbp]
    \centering
    \noindent\makebox[\textwidth]{\includegraphics[width=15cm]{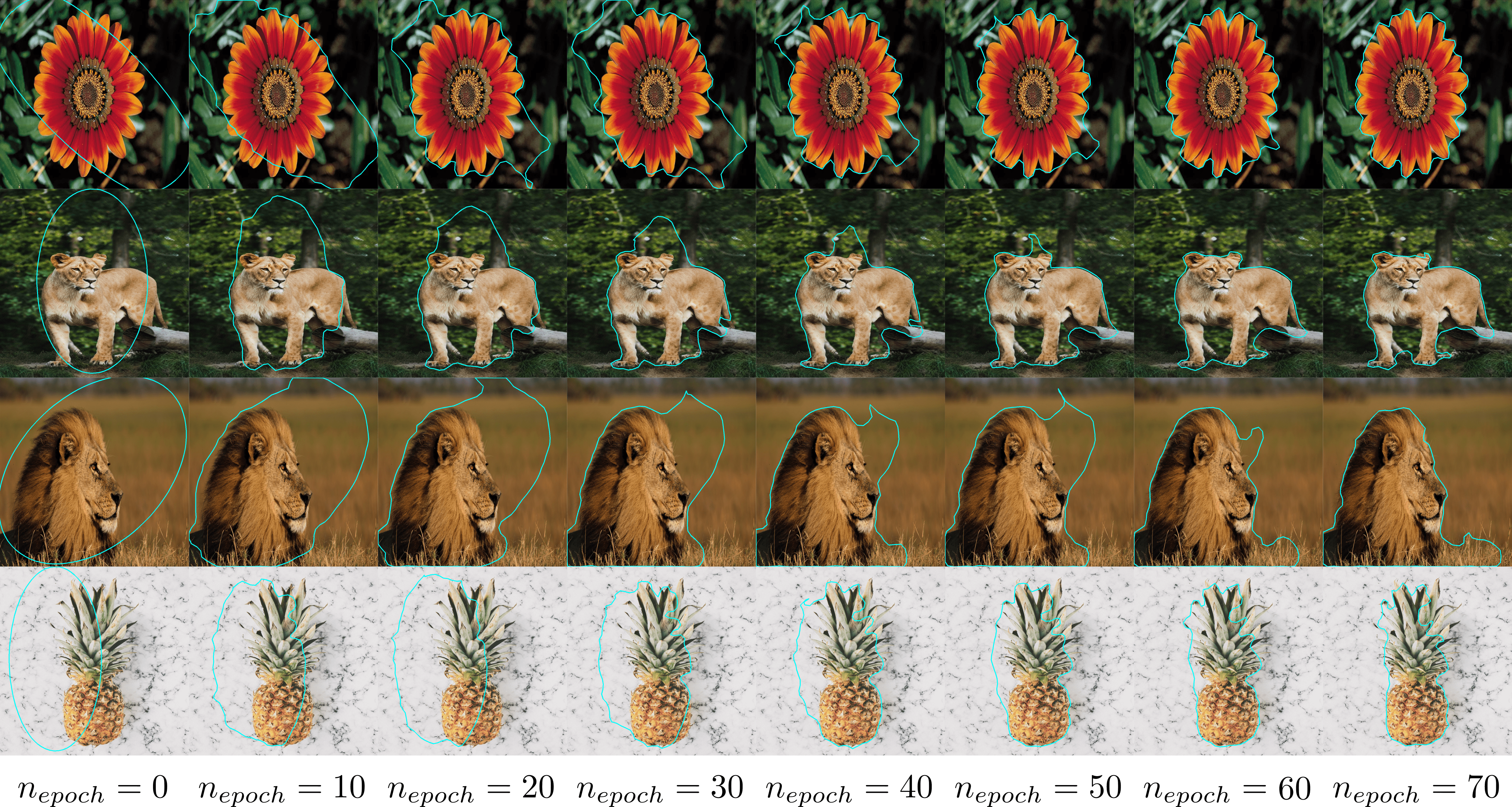}}
    \caption{Unsupervised DCF: evolution of the contour on four real-life images when varying the initial contour $C_0$.}
\label{unsupervised_images_flower}
\end{figure*}

\begin{figure*}[!htbp]
    \centering
    \noindent\makebox[\textwidth]{\includegraphics[width=15cm]{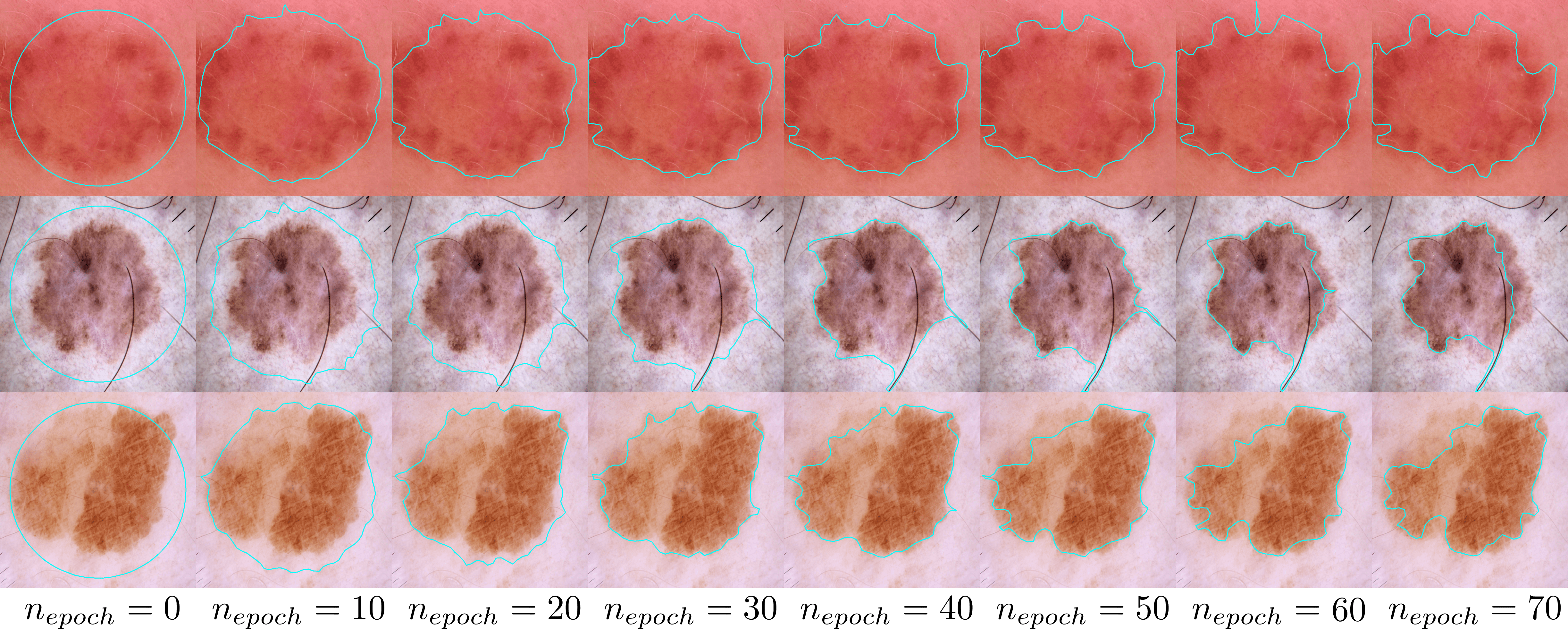}}
    \caption{Unsupervised DCF: evolution of the contour on three skin lesions from Skin Cancer MNIST: HAM10000 \cite{SKIN_LESION1, SKIN_LESION2}.}
\label{unsupervised_skin_lesion}
\end{figure*}

\begin{figure*}[!htbp]
    \centering
    \noindent\makebox[\textwidth]{\includegraphics[width=15cm]{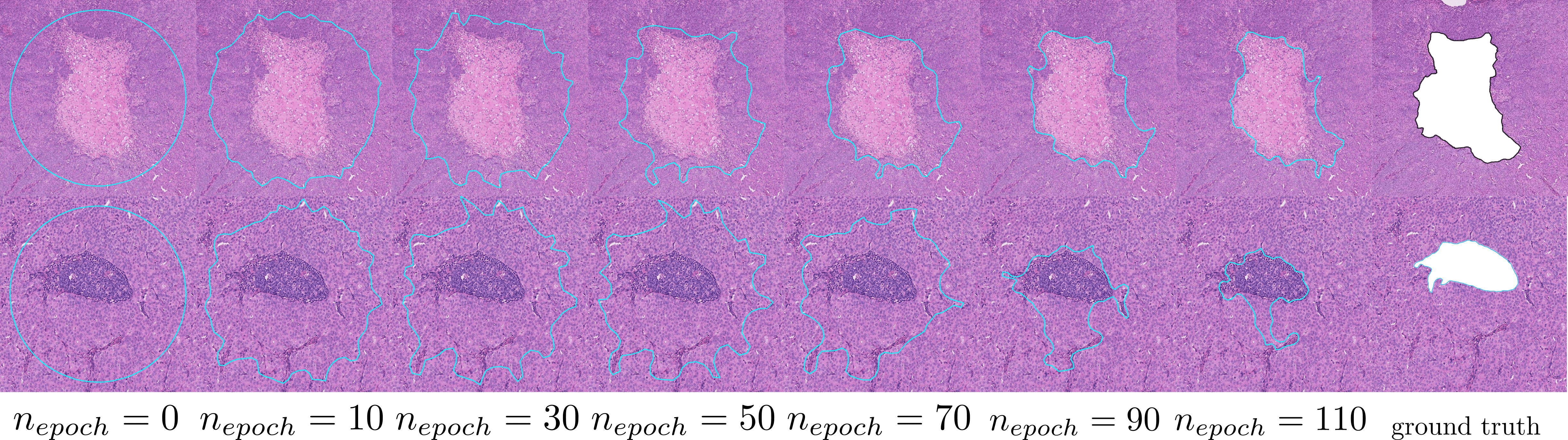}}
    \caption{Unsupervised DCF: evolution of the contour on two tumour areas from the CAMELYON16 \cite{CAMELYON} dataset.}
\label{unsupervised_images_tumor}
\end{figure*}

\renewcommand{\arraystretch}{1}

For the parametrization of the unsupervised DCF algorithm, we work with the parameters defined in \autoref{opti_param}.

\begin{table}[!htbp]
\centering
\caption{Parameters chosen for the unsupervised use cases.}

\begin{tabular}{p{11mm} p{7mm} p{7mm} p{5mm} p{5mm} p{7mm} p{6mm} p{4mm}}
\hline
                 & \boldmath$n_{nodes}$ & \boldmath$n_{epoch}$ & \boldmath$l_{r}$ & \boldmath$e_{d}$  & \boldmath$\lambda_{area}$ & \boldmath$t$\\ \hline
Histology  & 100       & 110         & $10^{-2}$   & 0.999 &  $5.$         & $10^{-2}$  \\ \hline
Real Life & 100       & 70         & $10^{-2}$          & 0.999   & 0.           & $10^{-2}$  \\ \hline
\end{tabular}
\label{opti_param}
\end{table}

\noindent  \autoref{unsupervised_images_flower} shows the evolution of the contour on four real-life images with different initial contours $C_0$. We can see that the model is robust to different initializations of the contour. The results remain consistent whether the object is completely within the initial contour or only partially. \autoref{unsupervised_skin_lesion} shows the evolution of the contour on three example skin lesions from Skin Cancer MNIST: HAM10000 \cite{SKIN_LESION1, SKIN_LESION2}. Finally, \autoref{unsupervised_images_tumor} shows the evolution of the contour on two histology images from the CAMELYON16 dataset \cite{CAMELYON}. The two segmented regions correspond to tumor areas in sentinel lymph nodes of breast cancer patients. These 2 last images are very different from the ImageNet dataset, yet the difference in texture is sufficient for the model to perform. However, in the second example the contour takes into account a small isolated tumor region that seems to have been overlooked in the ground truth. As the contour cannot divide itself, it leaves a poorly segmented area between the two regions.

The task becomes harder when working with histology images because the neural network is pre-trained on ImageNet \cite{IMAGENET}. As the features inside and outside of the contour are very similar, a local minimum may be reached when the contour completely collapse. To avoid this, we use the constraint on the area $\mathcal{F}_{\mathcal{A}rea}$ (see \autoref{area}) that prevents the contour from collapsing.\\
\indent In this section, we worked with real-life images of complex objects standing in front of rather homogeneous textured backgrounds and histology images showing tumors with homogeneous textures on a background also containing a homogeneous texture. As the VGG16 weights have been trained on ImageNet, it is not surprising that our method is able to segment objects with complex textures from real life. As for its use in histology, although the images are very different from real-life images, there is still sufficient texture contrast between foreground and background for our method to work.

On the other hand, if the proposed solution had to segment more complex objects with non-homogeneous textures in histology, the method as it stands would not work because it is purely driven by the texture difference between the inside and outside of the contour. So, in order to be able to perform such a task, we have also developed in the remainder of this work an approach requiring a single example of a non-homogeneous object in an histology image in order to successfully segment it on other images.

\section{Method: one-shot learning}
\label{oneshot}
\textcolor{black}{
In this section, we adapt DCF of \autoref{unsupervised} for one-shot learning to segment complex objects in histology such as  dilated tubules in kidney tissue. Dilated tubules possess a particular structure, characterized by a ring of nuclei around the periphery and a central lumen. This specific structure is ideal for working with the notion of distance map, which allows extracting isolines within an object of interest. The training dataset consists of a single support patch of a dilated tubule per WSI and its associated mask annotated by an expert. Given this single patch, we want to detect and segment all query dilated tubules in the entire WSI.}

\subsection{Fitting}
\label{fitting}
\textcolor{black}{Let $I^{sup} \in [0,1]^{H^{sup} \times W^{sup} \times 3}$ be the image of the dilated tubule and $M^{sup}\in \left\{ 0,1\right\} ^{H^{sup} \times W^{sup} \times 1}$ be the corresponding binary mask annotated by the expert. In the Fit and predict steps, we normalize the Whole Slide Images (WSIs) colors using the Macenko standard approach \cite{MACENKO} so that we remain robust to large staining differences. We work on instance segmentation for dilated tubules on Hematoxylin and Eosin $\text{H\&E}$ stained WSIs.}

\textcolor{black}{We first extract the VGG16 \cite{VGG16} features $\left\{f_s^{sup}, s \in \mathcal{S} = \left\{ 0,..,4 \right\} \right\}$ of $I^{sup}$ at each scale of the network. Then, we transform $M^{sup}$ into a normalized distance map $D^{sup} \in [0,1]^{H \times W \times 1}$ using Scikit-learn \cite{SCIKIT}. The normalized distance map indicates the normalized distance from the points inside of the contour to the contour itself. Hence, outside the contour the distance map is zero and inside, it is equal to the distance from the pixel to the nearest point on the contour divided by the maximum distance.}

\vspace{2mm}

\subsubsection{Isolines extraction}
\label{iso}
\textcolor{black}{Let $\mathcal{I}\subset[0,1]$ be the set of isoline normalized values. We first extract the isolines $\left\{Iso_i^{sup}, i \in  \mathcal{I} \right\}$.}
Theoretically, the isoline $Iso_i$ is defined as $Iso_i = \mathds{1}_{D^{sup} = i}$. However, the indicator is a non-differentiable function and we want to cover all the surface inside the mask with the extracted isolines. To do so we approximate the $\mathds{1}$ function with a decreasing exponential function. Then the isoline  becomes:
\color{black}
\begin{equation}
    \label{iso_gauss}
    Iso_i =  G(D^{sup}, i, \sigma)
\end{equation}

\textcolor{black}{
with $G(x,i, \sigma) = \exp{-\frac{(x - i)^{2}}{\sigma_i}}$ and  $\sigma$ chosen so that two consecutive isolines sum to $\frac{1}{2}$ at half distance of their centers.}

\vspace{4mm}
\subsubsection{Isoline features extraction}
\label{fis}
\textcolor{black}{We want to extract $(f_{i,s}^{sup})_{i \in \mathcal{I}, s \in \mathcal{S}}$, the features of the VGG16  for each scale $s$ and each isoline $Iso_i$. To compute $f_{i,s}^{sup}$ we downsample $Iso_i$ to  $(\frac{H^{sup}}{2^{s}},\frac{W^{sup}}{2^{s}})$, the spatial size of $f_s^{sup}$ with a bilinear interpolation to get $Iso_{i,s}$. To get the resulting $f_{i,s}^{sup}$ we compute the following:}
\color{black}
\begin{equation}
    \begin{split}
        f_{i,s}^{sup} & = L(Iso^{sup}_{i,s}, f_s^{sup})\\
    \end{split}
\label{L_second}
\end{equation}


\textcolor{black}{with $L$ defined in \autoref{definition_L}.}

\begin{figure}[!htbp]
  \centerline{\includegraphics[width = 9cm]{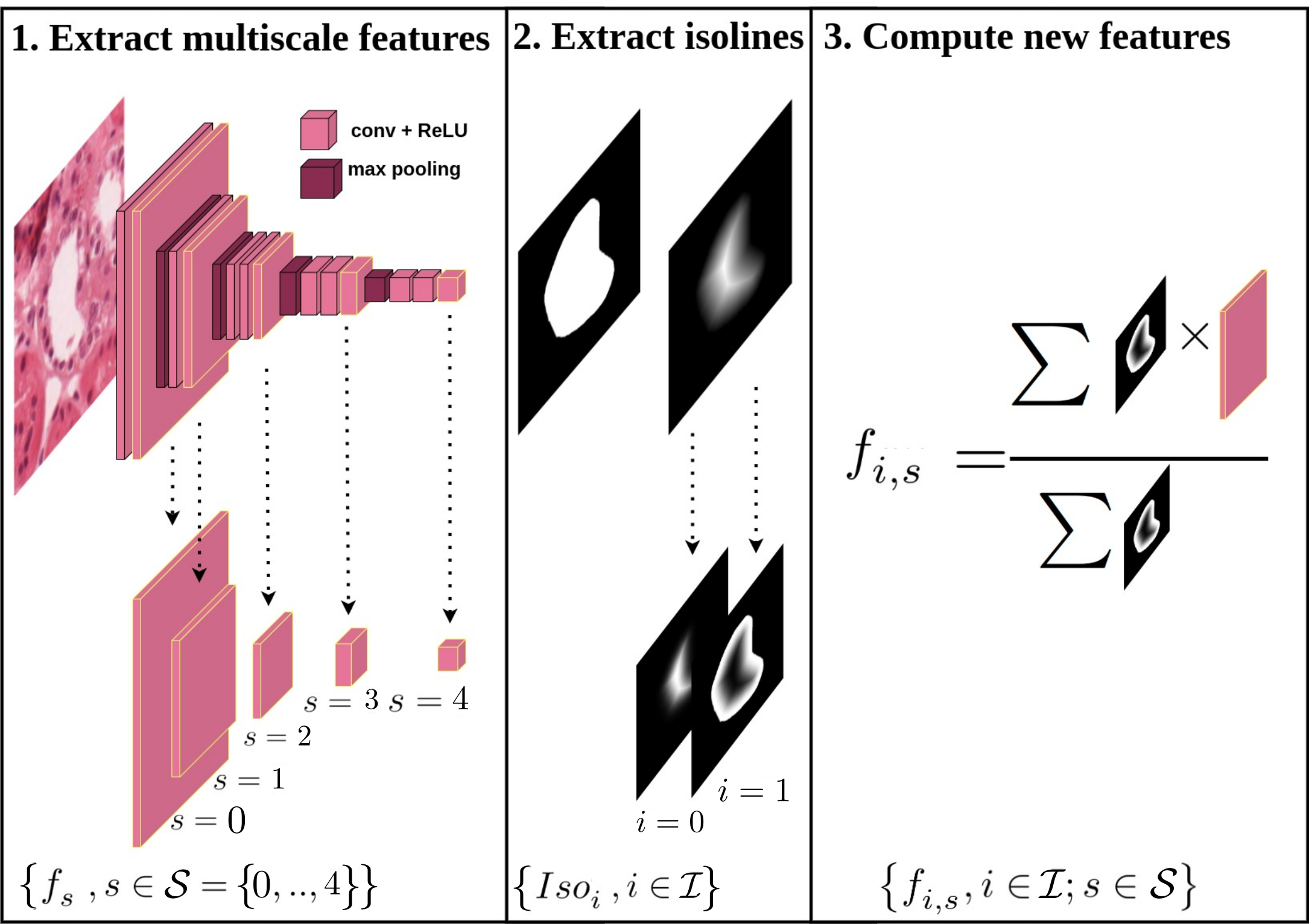}}
  \caption{Isoline features extraction pipeline. Two isolines are used in this example with $\mathcal{I} = \left\{ 0,1 \right\}$.}
 \label{isoline_features}
\end{figure}

\textcolor{black}{The collection of vectors $(f_{i,s}^{sup})_{i \in \mathcal{I}, s \in \mathcal{S}}$ aims to encode the support annotated by the expert based on the structural properties of dilated tubules. The overall isoline feature extraction process is illustrated in \autoref{isoline_features}. The extraction of  $(f_{i,s}^{sup})_{i \in \mathcal{I}, s \in \mathcal{S}}$ is done $n_{aug}$ times with a random augmentation applied to $(I^{sup}, M^{sup})$ at each time and then we compute the mean values. A random augmentation is done by applying basic augmentations sequentially with a given probability. We use  $90^{\circ}$ rotations, horizontal flip and vertical flip as basic augmentations with a probability $p=0.5$ for each of them.}

\subsection{Prediction}

\subsubsection{Extraction of all the potential dilated tubules}

\textcolor{black}{To segment the whole WSI, we need initial contours $C^0$ of candidate tubules. 
We use the prior knowledge that dilated tubules contain a white central lumen to extract all connected components (CCs) above the $90^{th}$ percentile of the grayscale values inside the tissue. Candidate query WSI patches are extracted around each selected connected component as the CC bounding box with some added margin. The corresponding initial contour for a patch is defined as the polygon nodes of the CC inside the patch. Each contour is then resampled as a polygon of $n_{nodes}$ equidistant nodes.}

\textcolor{black}{For the segmentation, we extract the same features $(f_{i,s}^{qu})_{i \in \mathcal{I}, s \in \mathcal{S}}$ of each query $I^{qu} \in [0,1]^{H^{qu} \times W^{qu} \times 3}$. Hence, we can minimize a cost function on the extracted features to evolve our contour $C$ with gradient descent. However, to do this, we need to create a fully differentiable function $F_{cd}$ that outputs a distance map from a contour $C$.}

\subsubsection{Fully differentiable contour-to-distance map function}
\label{dmap}

\textcolor{black}{
Using $F_{cm}$ defined in \autoref{contour_to_mask}, we define $F_ {cd}$ as:}
\color{black}
\begin{equation}
\label{contour_to_distance}
    \forall x \in \Omega, F_{cd}(x,C) = \frac{F_{cm}(x,C) \times d_2(x,C)}{max_{x \in \Omega}(F_{cm}(x,C) \times d_2(x,C))}
\end{equation}

\textcolor{black}{where, $d_2(x,C) = min_{i \in \left\{ 0,..,n_{nodes}\right\}} ||x - C_i||_2$}

\textcolor{black}{\autoref{increasing_k} (top row) shows the evolution of our approximation of the distance map when increasing $k$ and \autoref{increasing_nnodes} (bottom row) shows the evolution when increasing the number of nodes $n_{nodes}$.}

\vspace{3mm}
\subsubsection{Evolution of the contour}
\label{evo}

\textcolor{black}{In \autoref{mask} and \autoref{dmap} we built two differentiable functions. The first one, $F_{cm}$, turns a contour into a mask and the second one, $F_{cd}$, turns a contour into a distance map.}

\begin{figure*}[htbp!]
    \centering
    \noindent\makebox[\textwidth]{\includegraphics[width=16cm]{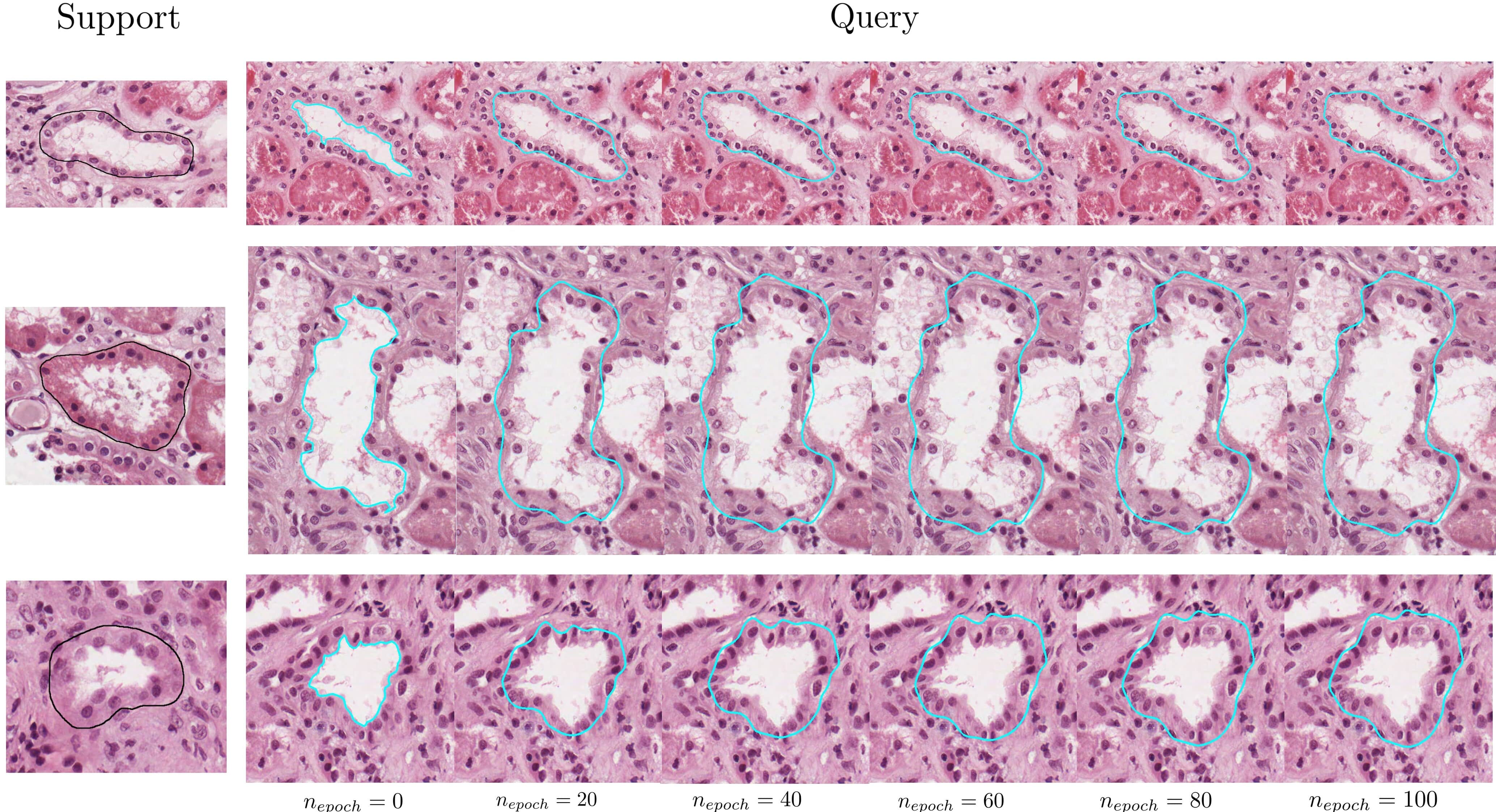}}
    \caption{Evolution of the contour with one-shot DCF. For each row, DCF is fitted on a "support" dilated tubule annotated by an expert (left part)  and the evolution of the contour on a "query" dilated tubule is shown on the right part.}
    \label{evo_tub}

\end{figure*}

\textcolor{black}{Evolution of the initial contour $C^0$ via loss minimization and gradient descent is described in  Algorithm \ref{dac}. While Algorithm \ref{dac_unsupervised} was designed for a single input image and compared features inside and outside the contour, Algorithm \ref{dac} takes 2 images as input and compares the features of the query with those of the support. Figure \ref{evo_tub} shows the evolution of the contour on 3 query tubule (right) using features extracted from 3 support tubule and their delimited mask (left).}

\begin{algorithm}[htb]
\caption{Deep ContourFLow}
\label{dac}
\begin{algorithmic}
\vspace{2.05mm}
\Algphasebegin{~~Phase 1 - Fit (one-shot learning on "support")}\\
\textbf{Input}: $I^{sup}$, $M^{sup}$
\textbf{Output}: $f^{sup}$

\vspace{1mm}
\State{$f_{i,s}^{sup} \gets 0$}
\For{$j \in 1, ..., n_{aug}$}
    \State{$(I^{sup}, M^{sup}) \gets Aug(I^{sup},M^{sup})$}
    \State{$(f_s^{sup})_{s \in \mathcal{S}} \gets \text{VGG16}(I^{sup})$}
    \State{$D^{sup} \gets  D_{map}(M^{sup})$ use $D_{map}$ of Sklearn \cite{SCIKIT}}
    \For{$i \in \mathcal{I}, s \in \mathcal{S}$}
    
        \State{$Iso_{i, s}^{sup} \gets Downscale(G(D^{sup}, i),scale = s)$ (\ref{iso_gauss})}
        \State{$f_{i,s}^{sup} \gets L(Iso_{i, s}^{sup}, f_s^{sup}) + f_{i,s}^{sup}$ (\ref{fis})}
    \EndFor
\EndFor
\State{$f^{sup} \gets f^{sup}/n_{aug}$}

\Algphaseinside{~~Phase 2 - Predict (Contour evolution on "query")}\\
\textbf{Input}: $(f_{i,s}^{sup})_{i \in \mathcal{I}, s \in \mathcal{S}}$, $I^{qu}$, $C^{0}$
\textbf{Output}: $C, score$
\State{$C \gets C^0$}
\State{$\Omega^{qu} \gets (i,j)_{i \in \llbracket 0,H^{qu}\rrbracket, j \in \llbracket 0,W^{qu}\rrbracket}$}
\State{$(f_s^{qu})_{s \in \mathcal{S}} \gets \text{VGG16}(I^{qu})$}
\State{$grad \gets +\infty$}
\For{$j \in 1, ..., n_{epochs}$}
    \If{$|grad| > t$}
        \State{$D^{qu} \gets F_{cd}(\Omega^{qu},C)$(\ref{contour_to_distance})}
        \For{$i \in \mathcal{I}, s \in \mathcal{S}$}
            \State{$Iso^{qu}_{i, s} \gets Downscale(G(D^{qu}, i),scale = s)$ (\ref{iso_gauss})}
            \State{$f^{qu}_{i,s} \gets L(Iso^{qu}_{i, s}, f_s^{qu})$ (\ref{L_second})}
        \EndFor
    \State{$Loss \gets loss_2(f^{sup},f^{qu})$ (\ref{loss})}
    \State{$grad \gets \nabla_C Loss$}
    \State{$C \gets C - l_{r} \times \text{Blur}(grad)$}
    \State{$C \gets$ Clean$(C)$}
    \State{$C \gets \text{Interp}(C)$}
    \EndIf
    \State{\textbf{else}{~Break$;$}}
\EndFor
\State{$score = \text{Sim}((f_s^{sup})_{s \in \mathcal{S}},(f_s^{qu})_{s \in \mathcal{S}})$}\\
\Return{$C, score$}

\end{algorithmic}
\end{algorithm}

\subsubsection{Loss and iteration adjustments}
\label{loss_it_adj}
As the loss function for the one-shot version, we use the following:

\begin{equation}
    \label{loss}
    loss_2(f^{sup}, f^{qu}) = \frac{1}{|\mathcal{S}||\mathcal{I}|}\sum_{s \in \mathcal{S}} \sum_{i \in \mathcal{I}} \frac{w_i}{2^s}\frac{||f_{i,s}^{sup} - f_{i,s}^{qu}||_2}{dim(f_{i,s}^{sup})}
\end{equation}

The purpose of this loss is to minimize the distance between the features extracted from the query and those of the support using a weighted $L_2$ norm $||\cdot ||_2$. As described in Algorithm \ref{dac} and displayed in \autoref{iterations_adjustments_images}, we also add a Gaussian blur $B$ on the gradient direction $\nabla_C loss_2$ to regularize the contour evolution. The operator $B$ prevents isolated points on the contour to push in a completely different direction from that of its neighbours.

\subsection{Deep ContourFLow: Classification}
Once the contour has stabilized, we now need to decide whether or not the object inside is a dilated tubule. To that end, we add a classification task to eliminate false tubule segmentation due to erroneous initial selection of query patches. For the classification task we threshold the following similarity metric:

\begin{align}
        \text{Sim}((f_s^{sup})_{s \in \mathcal{S}},(f_s^{qu})_{s \in \mathcal{S}}) = \sum_{s\in \mathcal{S}} \frac{1}{2^s} cos([f_s^{sup}]^{in},[f_s^{qu}]^{in})
\label{similarity_features}
\end{align}

This similarity is a weighted sum of cosine similarities between features spatially averaged at each scale of the support object and the query.

\section{Dataset and Results}

\subsection{Parameters}

For the parametrization of the algorithm, we work with $n_{nodes} = 100$, $k = 10^5$ and two isolines: $\mathcal{I} = \left\{0,1\right\}$. The isoline centered on 0 will handle the contour and the second one will handle the lumen of the dilated tubule. For the weighting of each isoline in \autoref{loss}, we use $w_0 = 0.1$ and $w_1 = 1 - w_0$. This stems from the fact that the central white area must stay in the center of the evolving contour and $C_0$ starts on the lumen and must stop at the edge of the dilated tubule. The tubule external edge must therefore be given greater focus. For augmentations, we use $n_{aug} = 100$. For the gradient descent, we use the following parameters: learning rate $l_{r} = 5 \times 10^{-2} $, with an exponential decay at $e_d = 0.999$ (default value)  a stopping criterion $t = 10^{-2}$ on the norm of the gradient and a maximum number of epochs $n_{epochs} = 300$. To compare the methods SGONE, CANET and DCF (\textit{Ours}), we used a scoring threshold maximising the F1 score for each one of them.

\subsection{Dataset}
\label{dataset}
\begin{figure}[htb]
  \centerline{\includegraphics[width = 8cm]{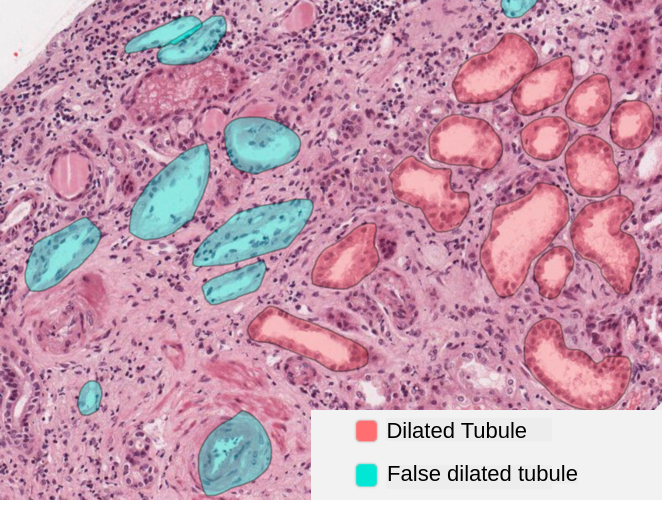}}
  \caption{Zoom on one of the fifteen WSIs of the annotated dataset, with false dilated tubules in turquoise and dilated tubules in red. }
 \label{dataset_image}
\end{figure}

\begin{figure}[htb]
  \centerline{\includegraphics[width = 8cm]{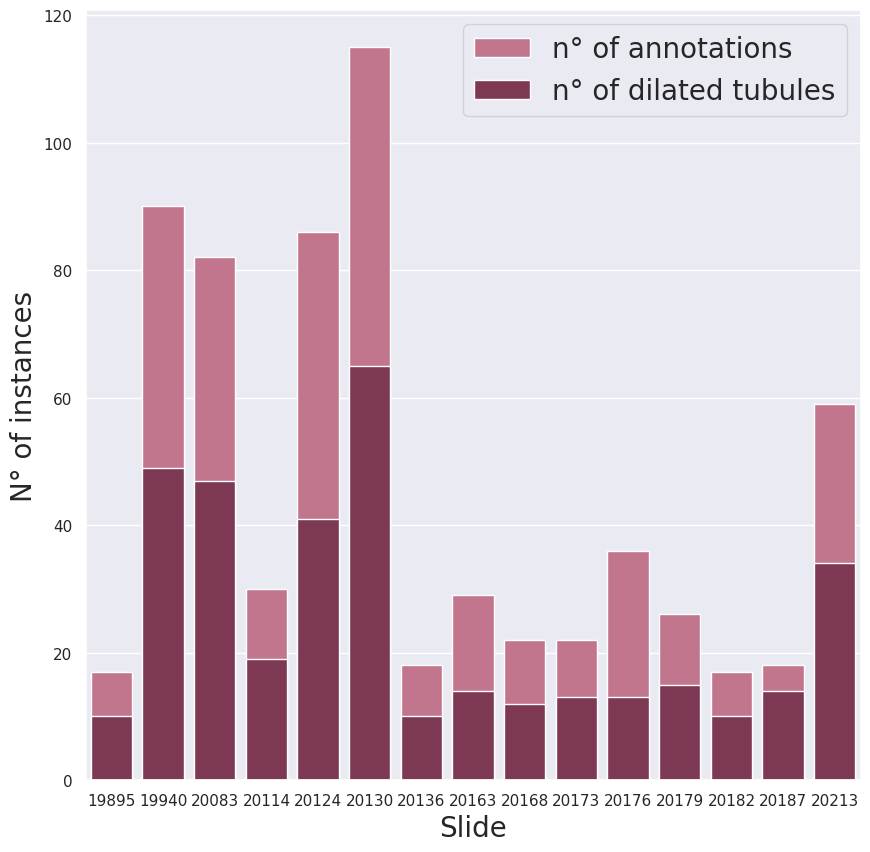}}
  \caption{Distribution of annotations in each WSI in the dataset. We can see that the dataset has been annotated to balance the two classes: false dilated tubule and dilated tubule.}
 \label{dataset_distribution}
\end{figure}

Fifteen $\text{H\&E}$ kidneys WSIs were used from the AIDPATH \cite{AIDPATH} kidney dataset. On each of these WSIs, we annotated dilated tubules but also "false dilated tubules" corresponding to white areas in the tissue that are not the lumens of dilated tubules (see \autoref{dataset_image}). The aim is to segment as accurately as possible, while at the same time classifying them properly. The total number of annotations contains 375 dilated tubules and  325 false dilated tubules for a total of 700 instances distributed over the fifteen WSIs. All annotations and codes are available at \url{https://github.com/antoinehabis/Deep-ContourFlow}. The distribution of annotations among the WSI used in the dataset is shown in \autoref{dataset_distribution}.

\subsection{Results}
\label{res}

\textcolor{black}{We tested our model for each WSI using one random annotated tubule as the support patch and predicted on all other query patches. We repeated the experiment ten times per WSI using ten different randomly chosen support patches to evaluate the robustness of the method. For the segmentation task, we report the Dice score on all True Positive dilated tubules. For the classification task, we report Precision, Recall, $F_1$ score. For both tasks, we report the Panoptic scores as in \cite{PANOPTIC} on all detected then segmented tubules. Results reported in \autoref{table_res} show for each model, the mean over all the WSI of the dataset of the max, mean, min and std of all the metrics for the ten experiments. Both CANET \cite{CANET} and SG-ONE \cite{SGONE} were trained on the PASCAL-5i dataset \cite{OSLSM}.
}
\begin{table}[!htbp]
\centering
\caption{Instance segmentation results of dilated tubules with DCF and CANET\cite{CANET}.}
\begin{center}
\scalebox{0.9}{
\begin{tabular}{l l|l|l|l|l|}
\cline{2-6}
\multicolumn{1}{l|}{} & Dice($\%)$             &   Precision($\%$)        & Recall($\%$)             & $F_1$($\%$)                & Panoptic($\%$)    \\ \hline
\multicolumn{1}{|l||}{\textbf{DCF}}   &                         &                          &                          &                          &                   \\ \hline
\multicolumn{1}{|l||}{\textit{mean}}                  & $\textbf{82.9}$ & $\textbf{69.8} $ & $\textbf{88.0}$ & $\textbf{76.1} $ & $\textbf{55.0} $ \\ \cline{2-6} 
\multicolumn{1}{|l||}{\textit{max}}                   & $87.2 $                 & $85.7$                   & $99.1$          & $86.4$                   & $64.9$\\ \cline{2-6} 
\multicolumn{1}{|l||}{\textit{min}}                   & $75.7  $                & $59.3$                   & $63.0$          & $64.3$                   & $43.6$\\ \cline{2-6} 
\multicolumn{1}{|l||}{\textit{std}}                   & $3.0$                   &$0.1$                     & $0.1$           & $0.1$                    & $6.5$                   \\ 

\hline

\multicolumn{1}{|l||}{\textbf{CANET\cite{CANET}}}   &                         &                          &                          &                          &                   \\ \hline
\multicolumn{1}{|l||}{\textit{mean}}                  & $51.9$ & $43.0$ & $41.9$ & $41.1$ & $15.6 $ \\ \cline{2-6} 
\multicolumn{1}{|l||}{\textit{max}}                   & $60.3 $                 & $54.4$                   & $62.4$          & $53.3$                   & $22.0$\\ \cline{2-6} 
\multicolumn{1}{|l||}{\textit{min}}                   & $32.9  $                & $27.0$                   & $24.9$          & $25.9$                   & $7.6$\\ \cline{2-6} 
\multicolumn{1}{|l||}{\textit{std}}                   & $8.3$                   &$9.1$                     & $11.4$           & $8.7$                    & $4.6$\\  \cline{2-6}
\hline
\label{table_res}

\end{tabular}}
\end{center}
\end{table}

\textcolor{black}{We can see that DCF outperforms CANET \cite{CANET} for all the metrics. SG-ONE \cite{SGONE} is unable to segment any tubules of the dataset so we did not mention it in \autoref{table_res}. Indeed, on histology data the latter has trouble generalizing the segmentation task using features learned from a dataset of real-life images. DCF obtains a relatively high Dice score. Detection scores are slightly lower, but this is logical as it is more difficult to recognize a dilated tubule from a false one using only raw ImageNet \cite{IMAGENET} features.}

\begin{figure}[!htbp]
    \centering
    \includegraphics[width=1\linewidth]{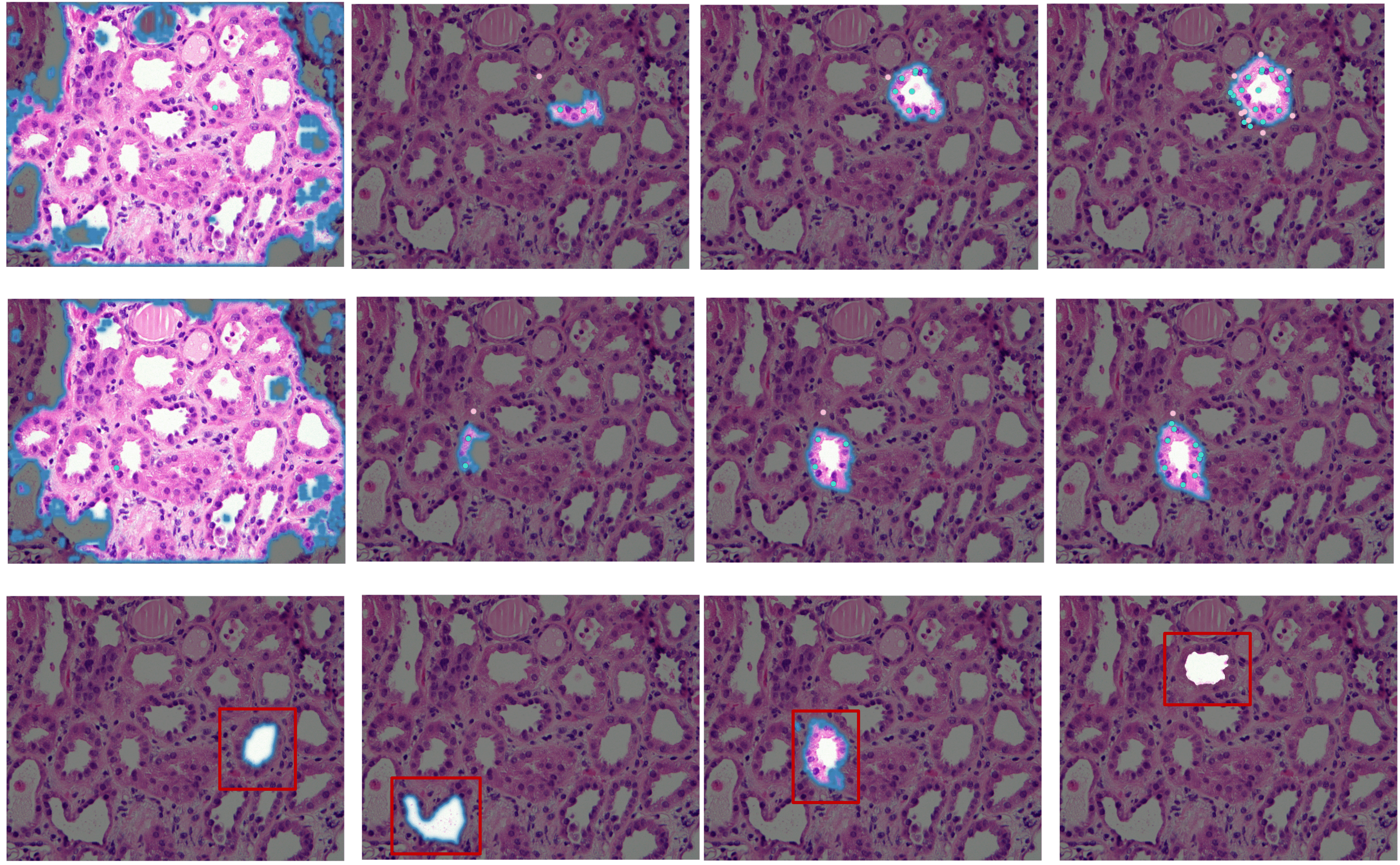}
    \caption{Results of Segment Anything (SAM) on dilated tubules. The first two rows show the resulting segmentation masks obtained with an increasing number of positive and negative clicks for two random dilated tubules. The last row shows the results of the segmentation using a bounding box on 4 different dilated tubules.}
    \label{sam_dilated}
\end{figure}

\textcolor{black}{We also show in \autoref{sam_dilated}, the results of the segmentation obtained using Segment Anything \cite{SAM} with different types of interactions. Segment Anything \cite{SAM} needs many interactions to begin to understand what a dilated tubule corresponds to (\autoref{sam_dilated} first two rows), which is impractical for pathologists. As for the use of bounding boxes, in most cases the algorithm is mistaken and only takes into account the lumen of the dilated tubule (\autoref{sam_dilated} last row). It appears that Segment Anything \cite{SAM} works very well on real-life images, but the algorithm does not perform as well when it comes to segment non-homogeneous objects of interest in histology.}

\section{Implementation details}
\textcolor{black}{In \autoref{unsupervised} and \autoref{oneshot}, we apply $F_{cm}$ in \autoref{contour_to_mask} and $F_{cd}$ in \autoref{contour_to_distance} on an image indexing grid of size $H \times W$. However, these functions have a fairly high computational cost because they require $\mathcal{O}(H \times W \times n_{nodes})$ operations. To reduce the complexity, we compute these functions with a mesh of size $\frac{H}{2^{2}} \times \frac{W}{2^{2}}$. It corresponds to the mask and the distance map obtained at scale $s=2$. Instead of calculating these values at scale $0$ and downsampling them to obtain the values at scales $1,2,3$ and $4$ we calculate the values at scale $s=2$, upsample to obtain those at scale $s=0,1$ and downsample to obtain those at scales $s=3,4$. This implementation reduces computation complexity by a factor of $2^2 \times 2^2  = 16$.}\\

\section{Limitations}
\textcolor{black}{
Our algorithm has two main limitations. First, its execution time is presently $\approx 1s$ per image. Even if reasonable, it could be greatly reduced by developing a batch-processing version allowing parallelization of tasks. The execution time also depends on the size of the image and the number of nodes in the polygon. Secondly, the polygon cannot be divided into several sub-polygons so it is only possible to segment one connected component at a time.}

\section{Conclusion}
\textcolor{black}{In this article, we propose a new framework, Deep ContourFlow (DCF), for both one-shot and unsupervised segmentation of complex objects in front of complex backgrounds. DCF can be applied to images of any size and does not require any training to be used. The unsupervised DCF is created to segment histology and real-life objects of interest in front of a textured background. Its advantage lies in its ability to consider both local and global information in images using deep learning features that facilitate the contour's evolution.}
\textcolor{black}{We also provide annotations of dilated tubules from the AIDPATH dataset \cite{AIDPATH}. From the results obtained, our algorithm demonstrates that it is possible to exclusively extract raw features using a VGG16 trained on ImageNet in order to segment histology images. The key to our one-shot algorithm lies in using the right projection operations on the raw VGG16 encoding of the support object along with using the appropriate distance measure in the projection space.}

\textcolor{black}{We compared our results to two state-of-the-art one-shot learning methods that have the a priori advantage of being fine tuned on PASCAL-5i dataset \cite{OSLSM} contrary to ours. However we show that these two methods have difficulties generalizing the features learned from \cite{OSLSM} to a histology dataset leading to poor results in the end.}

\section*{Acknowledgment}
\textcolor{black}{
This project has received funding from the Innovative Medicines Initiative 2 Joint Undertaking
under grant agreement No 945358. This Joint Undertaking receives support from the European Union’s Horizon 2020 research and innovation program and EFPIA. www.imi.europe.eu. The project was also partially funded through the PIA INCEPTION program (ANR-16-CONV-0005).
}

\section*{References and Footnotes}


\begin{thebibliography}{34}
\setcounter{enumiv}{0}
\color{black}
\bibitem{DEEP_SNAKE}{S. Peng, W. Jiang, H. Pi, X. Li, H. Bao, and X. Zhou, “Deep snake for real-time instance segmentation,” Proc. \emph{IEEE Comput. Soc. Conf. Comput. Vis. Pattern Recognit.}, pp. 8530–8539, 2020, doi: 10.1109/CVPR42600.2020.00856.}

\bibitem{POLY_RNN}{L. Castrejón, K. Kundu, R. Urtasun, and S. Fidler, “Annotating object instances with a polygon-RNN,” Proc. - \emph{30th IEEE Conf. Comput. Vis. Pattern Recognition, CVPR 2017}, vol. 2017-Janua, pp. 4485–4493, 2017, doi: 10.1109/CVPR.2017.477.}

\bibitem{POLY_RNN++}{D. Acuna, H. Ling, A. Kar, and S. Fidler, “Efficient Interactive Annotation of Segmentation Datasets with Polygon-RNN++,” Proc. \emph{IEEE Comput. Soc. Conf. Comput. Vis. Pattern Recognit.}, pp. 859–868, 2018, doi: 10.1109/CVPR.2018.00096.}

\bibitem{END}{H. Ali, S. Debleena and T. Demetri, “End-to-end trainable deep active contour models for automated image segmentation: Delineating buildings in aerial imagery,” in Computer Vision – ECCV 2020, Cham, 2020, pp. 730–746.
}

\bibitem{OTSU}{N. Otsu, "A Threshold Selection Method from Gray-Level Histograms," in \emph{IEEE Transactions on Systems, Man, and Cybernetics}, vol. 9, no. 1, pp. 62-66, Jan. 1979, doi: 10.1109/TSMC.1979.4310076.}

\bibitem{SNAKE}{M. Kass, A. Witkin, and D. Terzopoulos, “Snakes: Active contour models,” \emph{Int. J. Comput. Vis.}, vol. 1, no. 4, pp. 321–331, 1988, doi: 10.1007/BF00133570.}

\bibitem{GEODESIC}{R. Goldenberg, R. Kimmel, E. Rivlin, and M. Rudzsky, “Fast geodesic active contours,” \emph{Lect. Notes Comput. Sci.} (including Subser. Lect. Notes Artif. Intell. Lect. Notes Bioinformatics), vol. 1682, no. 1, pp. 34–45, 1999, doi: 10.1007/3-540-48236-9\_4.}



\bibitem{GVF}{C. Xu and J. L. Prince, “Generalized gradient vector flow external forces for active contours,” \emph{Signal Processing}, vol. 71, no. 2, pp. 131–139, 1998, doi: 10.1016/s0165-1684(98)00140-6.}

\bibitem{ACWE}{T. F. Chan and L. A. Vese, “Active contours without edges,” \emph{IEEE Trans. Image Process.}, vol. 10, no. 2, pp. 266–277, 2001, doi: 10.1109/83.902291.}

\bibitem{JC}{C. Zimmer and J. C. Olivo-Marin, “Coupled parametric active contours,” \emph{IEEE Trans. Pattern Anal. Mach. Intell.}, vol. 27, no. 11, pp. 1838–1842, 2005, doi: 10.1109/TPAMI.2005.214.}

\bibitem{JC2}{A. Dufour, V. Meas-Yedid, A. Grassart, and J. C. Olivo-Marin, “Automated quantification of cell endocytosis using active contours and wavelets,” Proc. \emph{Int. Conf. Pattern Recognit.}, pp. 25–28, 2008, doi: 10.1109/icpr.2008.4761748.}

\bibitem{GRAPHCUT}{Y. Y. Boykov, “Interactive Graph Cuts,” no. July, pp. 105–112, 2001.}

\bibitem{GRABCUT}{C. Rother, V. Kolmogorov, and A. Blake, “‘GrabCut,’” \emph{ACM Trans. Graph.}, vol. 23, no. 3, pp. 309–314, 2004, doi: 10.1145/1015706.1015720.}



\bibitem{ALEXNET}{A. Krizhevsky, I. Sutskever, and G. E. Hinton, “Imagenet classification with deep convolutional
neural networks,” in \emph{Advances in Neural Information
Processing Systems}, vol. 25.}

\bibitem{VGG16}{K. Simonyan and A. Zisserman, “Very deep convolutional networks for large-scale image recognition,” \emph{3rd Int. Conf. Learn. Represent. ICLR 2015} - Conf. Track Proc., pp. 1–14, 2015.}

\bibitem{INCEPTION}{C. Szegedy, V. Vanhoucke, S. Ioffe, J. Shlens, and Z. Wojna, “Rethinking the Inception Architecture for Computer Vision,” Proc. \emph{IEEE Comput. Soc. Conf. Comput. Vis. Pattern Recognit.}, vol. 2016-December, pp. 2818–2826, 2016, doi: 10.1109/CVPR.2016.308.}


\bibitem{GOOGLE_NET}{C. Szegedy et al., “Going deeper with convolutions,” Proc. \emph{IEEE Comput. Soc. Conf. Comput. Vis. Pattern Recognit.}, vol. 07-12-June-2015, pp. 1–9, 2015, doi: 10.1109/CVPR.2015.7298594.}

\bibitem{RESNET}{K. He, X. Zhang, S. Ren and J. Sun, "Deep Residual Learning for Image Recognition," arXiv, 2015, doi:10.48550/arXiv.1512.03385
}

\bibitem{USCF}{X. Zhou and N. L. Zhang, “Deep Clustering with Features from Self-Supervised Pretraining,” 2022, [Online]. Available: http://arxiv.org/abs/2207.13364.}


\bibitem{USCF2}{W. Kim, A. Kanezaki, and M. Tanaka, “Unsupervised Learning of Image Segmentation Based on Differentiable Feature Clustering,” \emph{IEEE Trans. Image Process.}, vol. 29, pp. 8055–8068, 2020, doi: 10.1109/TIP.2020.3011269.}

\bibitem{STYLEGAN_unsupervised}{R. Abdal, P. Zhu, N. J. Mitra, and P. Wonka, “Labels4Free: Unsupervised Segmentation using StyleGAN,” Proc. \emph{IEEE Int. Conf. Comput. Vis.}, pp. 13950–13959, 2021, doi: 10.1109/ICCV48922.2021.01371.}

\bibitem{VAE}{C. I. Bercea, B. Wiestler, and D. Rueckert, “{SPA} : Shape-Prior Variational Autoencoders for Unsupervised Brain Pathology Segmentation,” pp. 1–14, 2022.}

\bibitem{REVIEW}{N. Catalano and M. Matteucci, “Few Shot Semantic Segmentation: a review of methodologies and open challenges,” \emph{J. ACM}, vol. 1, no. 1, 2023, [Online]. Available: http://arxiv.org/abs/2304.05832.}

\bibitem{PFENET}{Z. Tian, H. Zhao, M. Shu, Z. Yang, R. Li, and J. Jia, “Prior Guided Feature Enrichment Network for Few-Shot Segmentation,” \emph{IEEE Trans. Pattern Anal. Mach. Intell.}, vol. 44, no. 2, pp. 1050–1065, 2022, doi: 10.1109/TPAMI.2020.3013717.}

\bibitem{SGONE}{X. Zhang, Y. Wei, Y. Yang, and T. S. Huang, “SG-One: Similarity Guidance Network for one-shot Semantic Segmentation,” \emph{IEEE Trans. Cybern.}, vol. 50, no. 9, pp. 3855–3865, 2020, doi: 10.1109/TCYB.2020.2992433.}

\bibitem{FSSPL}{N. Dong and E. P. Xing, “Few-shot semantic segmentation with prototype learning,” \emph{Br. Mach. Vis. Conf. 2018, BMVC 2018}, pp. 1–13, 2019.}

\bibitem{SMRCNN}C. Michaelis, I. Ustyuzhaninov, M. Bethge, and A. S. Ecker, “one-shot Instance Segmentation,” 2018, [Online]. Available: http://arxiv.org/abs/1811.11507.

\bibitem{OSLSM}{A. Shaban, S. Bansal, Z. Liu, I. Essa, and B. Boots, “One-shot learning for semantic segmentation,” \emph{Br. Mach. Vis. Conf. 2017, BMVC 2017}, 2017, doi: 10.5244/c.31.167.}

\bibitem{CNFSS}{K. Rakelly, E. Shelhamer, T. Darrell, A. Efros, and S. Levine, “Conditional networks for few-shot semantic segmentation,” \emph{6th Int. Conf. Learn. Represent. ICLR 2018} - Work. Track Proc., no. 2017, pp. 2016–2019, 2018.}

\bibitem{GANORCON}{O. Saha, Z. Cheng, and S. Maji, “GANORCON: Are Generative Models Useful for Few-shot Segmentation?,” Proc. \emph{IEEE Comput. Soc. Conf. Comput. Vis. Pattern Recognit.}, vol. 2022-June, pp. 9981–9990, 2022, doi: 10.1109/CVPR52688.2022.00975.
}

\bibitem{DATASETGAN}{Y. Zhang et al., “DatasetGAN: Efficient Labeled Data Factory with Minimal Human Effort,” Proc. \emph{IEEE Comput. Soc. Conf. Comput. Vis. Pattern Recognit.}, pp. 10140–10150, 2021, doi: 10.1109/CVPR46437.2021.01001.}



\bibitem{IMAGENET}{J. Deng, W. Dong, R. Socher, L.-J. Li, Kai Li, and Li Fei-Fei, “ImageNet: A large-scale hierarchical image database,” 2009 \emph{IEEE Conf. Comput. Vis. Pattern Recognit.}, pp. 248–255, 2010, doi: 10.1109/cvpr.2009.5206848.}


\bibitem{MUMFURD}{D. Mumford and J. Shah, “Optimal approximations by piecewise smooth functions and associated variational problems,” \emph{Commun. Pure Appl. Math.}, vol. 42, no. 5, pp. 577–685, 1989, doi: 10.1002/cpa.3160420503.}

\bibitem{BENTLEY}{J. L. Bentley and T. A. Ottmann, “Algorithms for Reporting and Counting Geometric Intersections,” \emph{IEEE Trans. Comput.}, vol. C–28, no. 9, pp. 643–647, 1979, doi: 10.1109/TC.1979.1675432.}

\bibitem{SKIN_LESION1}{N. Codella, V. Rotemberg, P. Tschandl, M. Emre Celebi, S. Dusza, D. Gutman, B. Helba, A. Kalloo, K. Liopyris, M. Marchetti, H. Kittler, A. Halpern: “Skin Lesion Analysis Toward Melanoma Detection 2018: A Challenge Hosted by the International Skin Imaging Collaboration (ISIC)”, 2018; https://arxiv.org/abs/1902.03368}

\bibitem{SKIN_LESION2}{P. Tschandl, C. Rosendahl,  \& H. Kittler, "The HAM10000 dataset, a large collection of multi-source dermatoscopic images of common pigmented skin lesions". \emph{Sci. Data 5}, 180161 doi:10.1038/sdata.2018.161 (2018).}


\bibitem{CAMELYON}{G. Litjens et al., “1399 H{$\&$}E-stained sentinel lymph node sections of breast cancer patients: The CAMELYON dataset,” \emph{Gigascience}, vol. 7, no. 6, pp. 1–8, 2018, doi: 10.1093/gigascience/giy065.}

\bibitem{MACENKO}{M. Macenko et al., “A method for normalizing histology slides for quantitative analysis,” Proc. - 2009 \emph{IEEE Int. Symp. Biomed. Imaging From Nano to Macro, ISBI 2009}, pp. 1107–1110, 2009, doi: 10.1109/ISBI.2009.5193250.}


\bibitem{SCIKIT}{F. Pedregosa and G. Varoquaux and A. Gramfort V. and V. Michel and B. Thirion and O. Grisel and M. Blondel and P. Prettenhofer and R. Weiss and V. Dubourg and J. Vanderplas and A. Passos and D. Cournapeau and M. Brucher and M. Perrot and E. Duchesnay, "Scikit-learn: Machine Learning in {P}ython,"JMLR.,vol. 12, pp. 2825-2830,2011}

\bibitem{AIDPATH}{G. Bueno, M. M. Fernandez-Carrobles, L. Gonzalez-Lopez, and O. Deniz, “Glomerulosclerosis identification in whole slide images using semantic segmentation,” \emph{Comput. Methods Programs Biomed.}, vol. 184, p. 105273, 2020, doi: 10.1016/j.cmpb.2019.105273.}

\bibitem{PANOPTIC}{A. Kirillov, K. He, R. Girshick, C. Rother, and P. Dollar, “Panoptic segmentation,” Proc. \emph{IEEE Comput. Soc. Conf. Comput. Vis. Pattern Recognit.}, vol. 2019-June, pp. 9396–9405, 2019, doi: 10.1109/CVPR.2019.00963.}

\bibitem{CANET}{C. Zhang, G. Lin, F. Liu, R. Yao, and C. Shen, “CANET: Class-agnostic segmentation networks with iterative refinement and attentive few-shot learning,” Proc. \emph{IEEE Comput. Soc. Conf. Comput. Vis. Pattern Recognit.}, vol. 2019-June, pp. 5212–5221, 2019, doi: 10.1109/CVPR.2019.00536.}

\bibitem{SAM}{A. Kirillov, E. Mintun, N. Ravi, S. Whitehead, A. C. Berg, and P. Doll, “Segment anything,” Proc. \emph{IEEE/CVF International Conference
on Computer Vision (ICCV)}, October 2023, pp. 4015–4026, doi: 10.1109/iccv51070.2023.00371.}
\end{thebibliography}
\end{document}